# AUTONOMOUS SURVEYING BOAT

Sajid Ullah[*1 2], Waseem Abdullah[1]

sajidullah.ist@gmail.com

[1] *Institute of Mechatronics Engineering, NWFP University of Engineering and Technology Peshawar*
[2] *Institute of Space and Technology, Islamabad Pakistan*

## *Abstract*

The fresh water reservoirs are one of the main power resources of Pakistan. These water reservoirs are in the form of Tarbela Dam, Mangla Dam, Bhasha Dam, Warsak Dam etc.[1]. To estimate the current power capability of the Dams, the statistical information about the water in the dam has to be clear. For the purpose of water management monthly or yearly survey of the dams required. One of the important parameter is to find the water level of the water, which can help us in finding the pressure and flow of water in dams. The existing surveying systems have some problems, i.e., risks, errors in measurement and sometimes expensive. Our project has tried a lot to overcome these flaws and to develop more economical, safe and accurate system for finding depth values of dams and ponds. The key purpose of Our Project "Autonomous Surveying Boat" is to have it log water depths along a predefined set of waypoints. The Autonomous Surveying Boat float in water according to predefined path, getting the coordinates from GPS Sensor and direction is controlled by using Magnetometer (compass) Sensor. It stores its data on SD card as a text file for later readings. The boat can also be used to find the average capacity of the dam. The average depth is calculated from the measured depth values at different set points of the dam. The actual length of the dam is determined by the magnetometer. The numbers of surveys over the time can help us in finding the silting ratio in dams. For square dams the length and width of the dam are measured and the average depth, then using these three parameters we can estimate the average capacity of the dam [2]. The boat is scalable for furthered modification if needed.
**Keywords :** Water level, Autonomous Surveying Boat, Average Depth,

## 1. INTRODUCTION

The objective of the paper is to implement an Unmanned Surface Vehicle (USVs) which measure water depth at predefined set points. It leads us to the silting level and area is also determined if it travels from one corner to another corner of the dam.It is more accurate, most economical and safe way of surveying the water in the dams. This project can be used for multipurpose, whether to find the depth of water, distance from one point to the other in water, the overall capacity of the dam or pond. Operating on the surface of water, USVs can be used for various missions that would be safer and cheaper than humans, such as marine environment monitoring, hydro logic survey, target object searching, and scientific study and so on [3]. While unmanned [water] surface vehicles (USV) date back at least to World War II, it is only in the 1990s that a large proliferation of projects appears, Corfield and Young (2006). This is in part due to the technological progress, but also driven by a paradigm shift of the US Navy with a much stronger focus on littoral warfare and anti-terrorism missions. Successful missions of USVs in the second Gulf war have increased interest within the US Navy in USVs and several modern navies followed suit. Potential USV missions could range from small torpedo-size data gatherers to large unmanned ships [4].

[*] *Corresponding Author*

Exploration of Remote environments, once the domain of intrepid Adventurers, can now be conducted in relative safety using unmanned vehicles [5]. And these unmanned vehicles if measure Depth values, actual distances of the path along which it travels, can enhance accuracy in the manual process Measuring depth by cable. And minimize manual labor and time. The flat-bottomed boat is used as surface vehicle. The GPS point can be taken either from bringing GPS receiver to the point of interest or from Google earth. The magnetometer is used to drive the boat in Straight line. The point where we find depth values are predefined Points for our controller. The depth values are acquired by the ultrasonic Ranging sensor. SD card module saves these depth values of the set points.

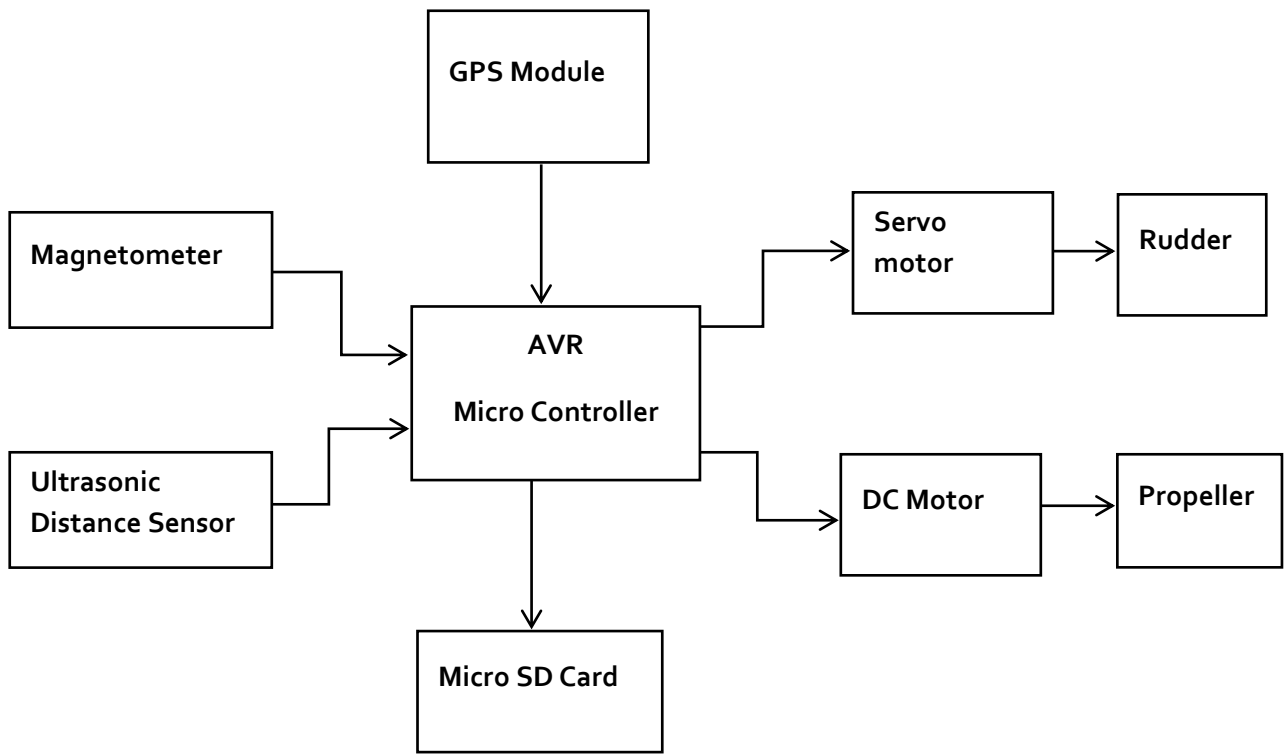

Fig 1 Block Diagram of Autonomous Surveying Boat

## 2. EXISTING SYSTEM

The existing system consists of many methods to find water depth.

One way to determine dam depth is to row out into the dam and lower a weighted line over the side. When the line is vertical, measure the length of the line needed to reach the bottom. Alternatively, use a pole with distances marked on it. You will need to do this at a number of places across the dam to find the deepest point [6].

An alternative for smaller dams, or if no boat is available, use a fishing line with a sinker on the bottom with a float attached. The line is cast out repeatedly, with the float gradually adjusted until it's not quite floating on the surface. The distance between the float and the sinker will be the depth at that point in the dam. Again, you will need to do this at a number of places across the dam.

For small dams measurement tap with a weight at one end is used to find the depth of water [7].

# 3. PROPOSED SYSTEM

The small size flat-bottom boat is used as an Autonomous surface vehicle. The GPS point can be taken either from bringing the GPS receiver to the point of interest, or from Google earth [8]. The magnetometer is used to drive the boat in required direction [9]. The point where we find depth values are predefined Points for our controller. The depth values are acquired by the ultrasonic Ranging sensor [10]. SD card module saves these depth values of the set points [11].

*3.1. BOAT MECHANICAL DESIGN*

The autonomous Surveying Boat is 18 x 12 inches in dimension, total costs $212.5 and having a weight of 5615 grams. All the circuits, actuators and sensors are mounted on the top of the boat, the Ultrasonic sensor is attached to the front side of boat in vertical position.

The boat hull was made from four PVC pipes. The pipes were 12 inches long and 4 inches in diameter

Table 1 Gross weight of Boat

| S.NO | COMPONENTS | WEIGHT(GRAMS) |
|---|---|---|
| 1 | 12v DC-battery | 230 |
| 2 | RadarSonics Model250 Sonar | 700 |
| 3 | Magnetometer ( | 4 |
| 4 | Johnson Electric 12V DC Power Motor | 149.6 |
| 5 | Servo motor | 37.2 |
| 6 | GPS module | 9 |
| 7 | Propeller | 25 |
| 8 | Rudder | 40 |
| 9 | Circuit boards | 30 |
| 10 | PVC pipes | 320 |
| 11 | Plastic thermo Foil | 50 |
| 12 | Aluminum Foil | 3800 |
| 13 | Cardboard | 80 |
| 14 | Miscellaneous | 140 |
| *Net Weight* | | 5615 |

*3.2. CIRCUITS*

3.2.1. H bridge

This is H Bridge IC L298N which gives a maximum of 2 ampere on its output pin, while the DC motor (propeller) draws a current of 1.1 ampere. It is used to control the rotation of the propeller.

*3.2.2. MAIN POWER BOARD*

It provides a voltage of 5 volts to each sensor, and also provides 9 volts to Arduino Board.

*3.3. SENSORS*

The Autonomous Surveying Boat uses three sensors for its operation. The Ultrasonic Sensor RadarSonics Model250 Sonar is used to find the depth. The GPS ReceiverSkylab SKM53 is used to determine the GPS coordinates when Autonomous Surveying Boat is floating in water to find the depth at predefined set points.The Honeywell HMC5883L is used as compass sensor.

Table 2 sensors

| Sensor | Variable | Range | Accuracy |
|---|---|---|---|
| *RadarSonics Model250 Sonar* | *Depth* | *0.4– 135 m* | *0.1m* |
| *HMC388L (Compass)* | *Heading, distance* | *$360^0$, $\pm 60^0$* | *$1^0$ to $2^0$ heading* |
| *SKM53 (GPS Receiver)* | *Time, position, track* | *-* | *4.5 – 10 meter* |

# 4. TESTS & RESULTS

As the main objective was to measure thedepth of the water, so first of all we started testing Ultrasonic Sensor. We have four pins in our Ultrasonic sensor RadarSonics Model250 Sonar . We programmed it in Arduino IDE using C language. In the program we put codes for observing depth values in Arduino serial monitor, so we observed the sensor output by changing the level of sensor over the water.

We have used SKYLAB SKM53 GPS Module for position coordinates. During testing, we note that the GPS receiver changes the last four digits after 4 meters in any direction.

After all these individual tests we interfaced all the sensors and actuators with our Microcontroller, and interconnected all the circuits. Our choice of location for testing our project is Swan Swimming pool which is located in Hayatabad. The Task was to find the depth of the water at specific GPS points. The "Autonomous Surveying Boat" follows a predefined in the water to collect depth points. The direction and turns guidelines are provided by the compass (Magnetometer). We tested different points by approaching through different paths.

In one of our test we collect three depth points of water at three GPS points. The points are stored in the SD card in the text file. The depth values are 17cm, 17cm and 18 cm. In our 2$^{nd}$ test we change the path and GPS points; this time the values were 17cm 16cm 17 cm.

For third test we choose more deeper and large pool from the previous one. This time found six depth points at different set points. The output of the sensor was 312cm, 317 cm, 311cm, 374cm, 382cm, and 380cm.

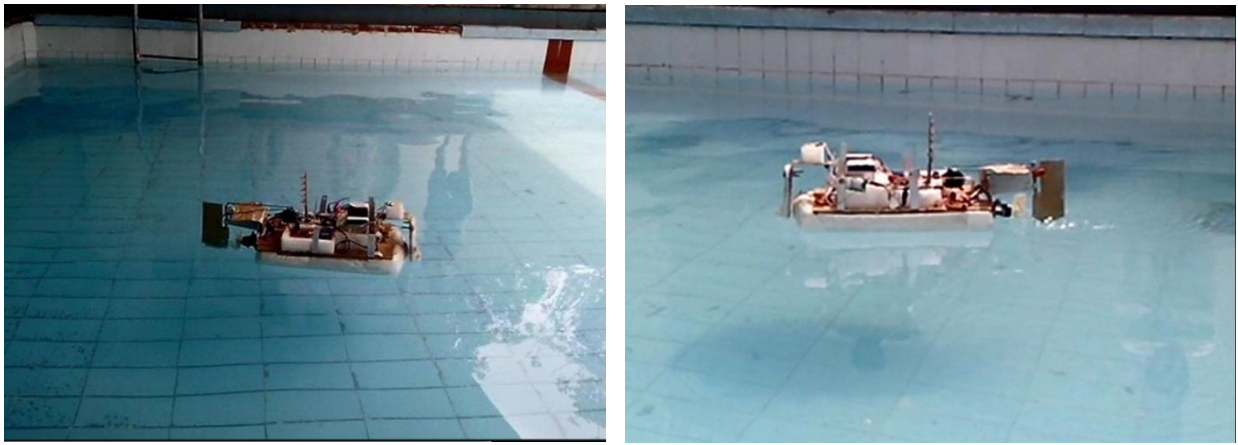

Figure 2 The Autonomous Surveying Boat on its way to find depth in first test

The following table consists of complete information of these 3 tests:

Table 3 Test No. 3 Data

| TEST# | Setpoint# | GPS SET POINTS | | DEPTH (cm) | Actual distance from start upto this point (m) |
|---|---|---|---|---|---|
| | | *LAT* | *LONG* | | |
| Test 1 | 1 | 33.971659 | 71.442073 | 17 | 4.50 |
| | 2 | 33.971648 | 71.442103 | 17 | 9.04 |
| | 3 | 33.971642 | 71.442063 | 18 | 13.64 |
| Test 2 | 1 | 33.971648 | 71.442124 | 17 | 4.52 |
| | 2 | 33.971670 | 71.442079 | 16 | 9.10 |
| | 3 | 33.971639 | 71.442073 | 17 | 14.07 |
| Test 3 | 1 | 33.971902 | 71.441588 | 346 | 5.01 |
| | 2 | 33.971880 | 71.441690 | 345 | 10.25 |
| | 3 | 33.971839 | 71.441583 | 359 | 15.52 |
| | 4 | 33.971777 | 71.441714 | 374 | 22.67 |
| | 5 | 33.971693 | 71.441588 | 382 | 30.01 |
| | 6 | 33.971688 | 71.441752 | 380 | 36.94 |

The set points at which depth is measured during the test 3 are marked with blue marks in the Google map as shown below.

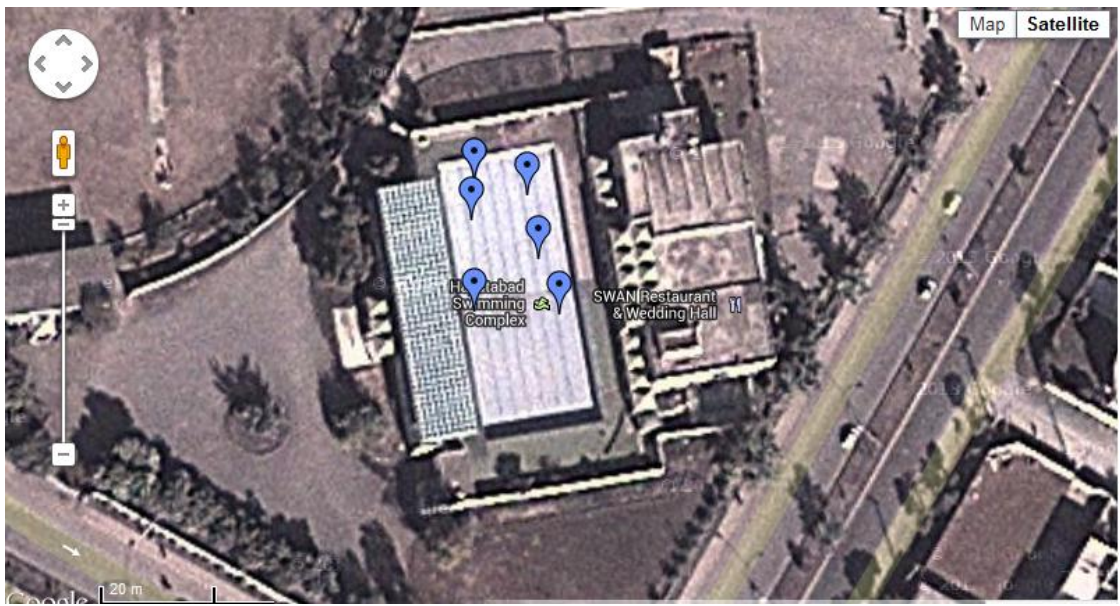

Fig3: Test Set points in Google Map

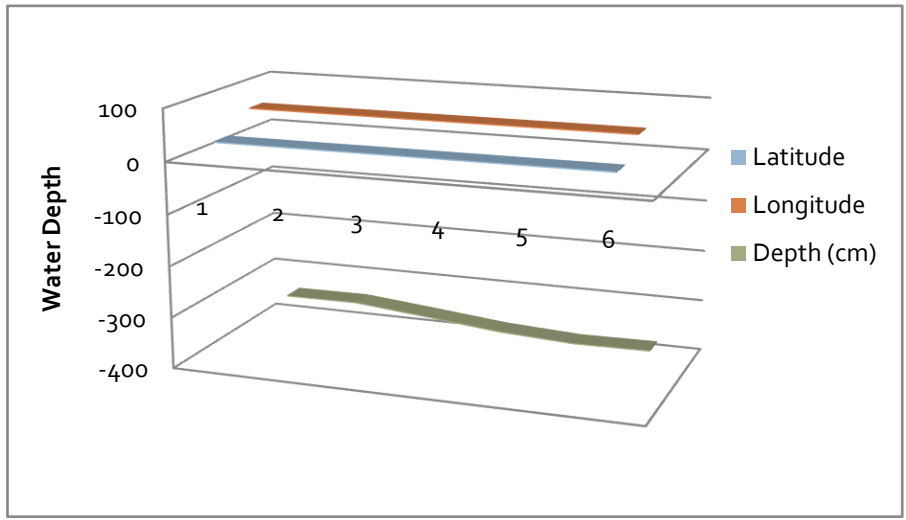

Figure 4 Depth Graph along Setpoints

## 5. CONCLUSION

An Autonomous Surveying Boat is designed and developed that can be used for finding the depth of water (in dams & ponds) and other related applications.The Boat is made to work autonomously according to predefined program.The boat is an example of onboard electronics requiring low electrical power, which provide some level of intelligence, surveillance, safest and economical method to find water depth. The robot is used keeping an eye on silting of dam, which proves to be inexpensive technique. Further computation is necessary to the mapping of bed of dam or other water reservoir.

## 6. ACKNOWLEDGMENT

The author gratefully acknowledge the help received from the institute of Mechatronics engineering,
University of Engineering and Technology Peshawar  and Management at Swan pool